\documentclass[11pt]{article}

\usepackage[preprint]{acl}

\usepackage{times}
\usepackage{latexsym}

\usepackage[T1]{fontenc}

\usepackage[utf8]{inputenc}
\usepackage{colortbl}
\usepackage{multicol}
\usepackage{multirow}
\usepackage{booktabs}
\usepackage{microtype}
\usepackage{algorithmicx}
\usepackage{algorithm}
\usepackage{algpseudocode}
\usepackage{amsmath}
\usepackage{inconsolata}
\usepackage{amssymb}

\usepackage{graphicx}

\title{PRISM-MCTS: Learning from Reasoning Trajectories with Metacognitive Reflection}

\author{
Siyuan Cheng, Bozhong Tian, YanChao Hao, Zheng Wei\\
  Tencent PCG \\
  \texttt{\{chancecheng,cobbtian,marshao,hemingwei\}@tencent.com}
}

\begin{document}
\maketitle
\newcommand{\our}{PRISM-MCTS}
\newcommand{\qwen}{Qwen3-30B-A3B-Instruct-2507}

\begin{abstract}
The emergence of reasoning models, exemplified by OpenAI-o1, signifies a transition from intuitive to deliberative cognition, effectively reorienting the scaling laws from pre-training paradigms toward test-time computation.
Though Monte Carlo Tree Search (MCTS) is promising in this domain, existing methods are often inefficient, treating rollouts as isolated trajectories and causing significant computational redundancy.
To address these limitations, we propose {\our}, a novel reasoning framework that draws inspiration from human parallel thinking and reflective processes.
{\our} integrates a Process Reward Model (PRM) with a dynamic shared memory, capturing both \textit{Heuristics} and \textit{Fallacies}.
By reinforcing successful strategies and pruning error-prone branches, {\our} effectively achieves refinement.
Furthermore, we develop a data-efficient training strategy for the PRM, achieving high-fidelity evaluation under a few-shot regime.
Empirical evaluations across diverse reasoning benchmarks substantiate the efficacy of {\our}. 
Notably, it halves the trajectory requirements on GPQA while surpassing MCTS-RAG and Search-o1, demonstrating that it scales inference by reasoning judiciously rather than exhaustively.
\end{abstract}
\section{Introduction}

\begin{figure}[htbp]
    \centering
    \includegraphics[width=0.47\textwidth]{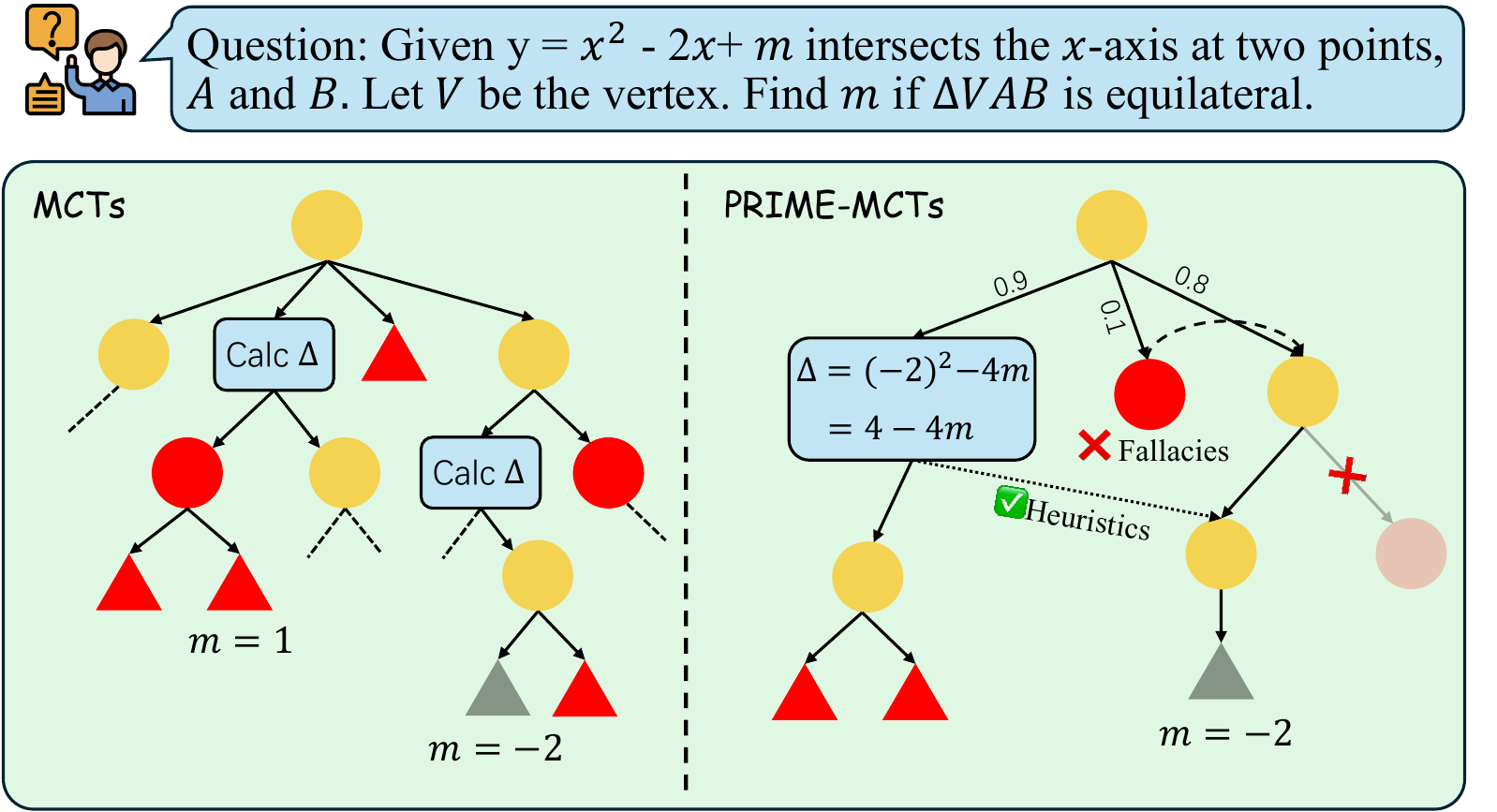}
    \caption{Standard MCTS (left) vs. {\our} (right). {\our} enables global information sharing to prune incorrect paths and reuse verified steps for higher efficiency.}
    \label{fig:intro}
\end{figure}

In recent years, Large Language Models (LLMs)~\cite{LLM_Survey, LLM_Survey2} have rapidly become a cornerstone in natural language processing, and the field of LLMs has been undergoing a paradigm shift.
Conventional LLMs, such as GPT~\cite{GPT4}, LLaMA~\cite{LLaMA}, and Qwen~\cite{Yang2024Qwen2_5}, primarily rely on a ``fast thinking'' reasoning paradigm.
This paradigm is primarily anchored in the knowledge and reasoning priors synthesized during pre-training, facilitating rapid and intuitive generation from a single query.
While ``fast thinking'' excels in routine tasks, it frequently falls short when addressing complex problems that necessitate multi-step iteration, rigorous logic, or dynamic adaptation to novel information.
With the emergence of models like OpenAI-o1~\cite{OpenAI_O1}, ``slow thinking'' has become a pivotal direction for enhancing reasoning~\cite{wei2022chain, Yao2023ToT, system2_survey}, simulating human deliberation by scaling up inference-time compute.

Among the various pathways to achieving ``slow thinking'' reasoning techniques based on MCTS have become prominent~\cite{Hu2025MCTsRAG, Qi2024rstar, zhang2024restmcts}. 
By modeling reasoning as a structured tree search with backtracking, MCTS boosts complex logical task performance~\cite{Chen2024AlphaMath}.
Nevertheless, these approaches are encumbered by prohibitive computational overhead stemming from exhaustive exploration, while their reliance on a sequential paradigm induces informational isolation and computational waste through redundant node re-simulations.
In resource-constrained or time-critical scenarios, balancing MCTS efficiency with reasoning quality is the critical bottleneck for ``slow thinking'' techniques.

To address these limitations, we propose \textbf{P}rocess-\textbf{R}ewarded \textbf{I}ntelligent \textbf{S}hared \textbf{M}emory MCTS ({\our}), as illustrated in Figure~\ref{fig:intro}.
Drawing from human metacognitive reflection, {\our} captures the interconnected nature of complex reasoning trajectories, leveraging accumulated experience to distill valuable guidance from even erroneous paths.
Consequently, these distilled insights effectively optimize the navigation of both concurrent and subsequent search trajectories.
Specifically, {\our} utilizes a PRM to conduct granular assessments of each reasoning step, subsequently isolating nodes with divergent rewards into a dynamic memory for classification as heuristics or fallacies.
During concurrent reasoning, knowledge from memory nodes is shared globally to foster system-wide synergy.
Specifically, \textit{Heuristics Memory} empowers subsequent nodes to exploit verified logic from antecedent trajectories to accelerate convergence, whereas \textit{Fallacies Memory} serves as a proactive constraints that prevent the redundant expenditure of computational resources on analogous errors.
Moreover, we employ a Memory Manager to efficiently distill key information from the repository's state.
Through these mechanisms, {\our} shifts the traditional serial reasoning paradigm toward a parallelized search framework supported by global information sharing.
Grounded in reflection and experiential sharing, this mechanism refines MCTS selectivity, thereby bolstering both reasoning efficiency and deductive rigor in intricate tasks.

In summary, the main contributions of our work are as follows:
\begin{itemize}
    \item We introduce {\our}, a parallelized reasoning framework that leverages a reflective memory mechanism to share global heuristics and prune fallacies, thereby optimizing search efficiency.
    \item We devise a dual-stage, few-shot PRM training method synergizing Step-wise DPO and multi-class classification for precise, granular process evaluation.
    \item Extensive experiments demonstrate that {\our} outperforms state-of-the-art baselines on complex reasoning tasks, achieving higher accuracy with significantly fewer search steps.
\end{itemize}
\section{Related Work}

\paragraph{Inference-time Scaling.}
Scaling inference-time computation has emerged as a compelling paradigm for enhancing the reasoning capabilities of LLMs \cite{openai2024reasoning, guo2025deepseek, hao2023reasoning, lightman2023let}.
Early efforts like Chain-of-Thought (CoT) prompting~\cite{wei2022chain} focused on linear trajectories, encouraging models to generate intermediate reasoning steps.
To improve the robustness of linear reasoning, Self-Consistency (CoT-SC)~\cite{wang2023cotsc} samples diverse reasoning paths and selects the final answer via majority voting.
Additionally, Program-of-Thought (PoT)~\cite{Chen2023PoT} generates executable code to offload computational sub-tasks to external interpreters.
However, these linear methods suffer from error propagation and limited flexibility.
To overcome these limitations, tree-based planning methods like Tree-of-Thought (ToT)~\cite{Yao2023ToT}, Graph-of-Thought (GoT)~\cite{besta2024graph}, and Monte Carlo Tree Search (MCTS) enable structured search with backtracking and evaluation, significantly improving complex reasoning~\cite{zhou2023language, gan2025master, Qi2024rstar}.

\paragraph{MCTS with Process Reward Models.}
MCTS provides a principled framework for balancing exploration and exploitation, demonstrating remarkable efficacy in domains requiring rigorous logic, such as mathematics and code generation~\cite{Zhou2024LATS, Zhang2023PGTD, Chen2024AlphaMath, xu2025sramcts}.
However, terminal outcome supervision causes inefficient exploration, as MCTS lacks intermediate guidance in vast search spaces.
To address this, recent studies have integrated PRMs directly into the MCTS framework to provide dense, step-level guidance.
By leveraging granular feedback to reduce error propagation, methods such as ReST-MCTS$^{*}$ \cite{zhang2024restmcts}, AR-MCTS~\cite{zhang2025ARMCTS}, and I-MCTS \cite{liang2025imcts} significantly enhance the effectiveness of the search process.
However, training effective PRM typically necessitates extensive supervision, posing significant challenges for data-scarce domains~\cite{Lightman2024prm, Uesato2022prm, zhang2025qwenprm}.
In this work, we propose a dual-stage, data-efficient training strategy tailored for MCTS-guided reasoning.
By combining step-wise preference learning with fine-grained value classification, {\our} enables precise few-shot evaluation, enhancing guided search scalability and effectiveness.

\paragraph{Reflective and Memory-Augmented Reasoning.}
Integrating memory mechanisms enables LLMs to transcend the limitations of static parametric knowledge, facilitating context retention and iterative refinement~\cite{shinn2023reflexion, packer2023memgpt, fang2025lightmem}.
While early approaches like Retrieval-Augmented Generation (RAG)~\cite{lewis2020retrieval} and IRCoT~\cite{Trivedi2023ircot} primarily focused on enhancing factual accuracy via external corpora, they often function as static lookups lacking dynamic adaptation during reasoning.
To enhance structured reasoning, recent studies have begun to incorporate these reflective and memory capabilities directly into MCTS frameworks.
For instance, MC-DML~\cite{Shi2025MCDML} and I-MCTS~\cite{liang2025imcts} employ introspection to refine intermediate steps, while CoAT~\cite{zhang2025coat} leverages associative memory to link related concepts during inference.
However, these methods typically aggregate experiences indiscriminately, failing to explicitly distinguish between valid logical patterns and distinct failure modes, which can lead to error recurrence.
In this work, we propose a structured dual-memory mechanism—\textit{Heuristics Memory} for reinforcing successful strategies and \textit{Fallacies Memory} for intercepting erroneous paths.
Governed by a \textit{Memory Manager}, this architecture dynamically filters high-fidelity insights to proactively prune redundant branches, transforming the search into a self-improving reasoning system.

\section{\our}
\label{sec:method}

\subsection{Preliminaries}

\paragraph{Monte Carlo Tree Search.}
MCTS is a widely used sampling-based search method for decision-making optimization. 
It constructs a search tree by repeatedly executing four steps: selection, expansion, simulation, and back-propagation.
During the selection phase, MCTS recursively selects child nodes from the root using the Upper Confidence Bounds applied to Trees (UCT)~\cite{Kocsis2006uct}.
The UCT of a node $n$ is calculated as follows:
\begin{equation}
\label{eq:uct}
    UCT(n) = V(n) + \epsilon\sqrt{\frac{lnN(p)}{N(n)}}.
\end{equation}
where $N(n)$ is the number of visits to node $n$, $V(n)$ is the score value, and $p$ is the parent node of node $n$. $\epsilon$ is the exploration weight. 
Upon episode completion, a back-propagation is performed to update the value of node $n$ and its parent nodes.

\begin{algorithm}[t]
\caption{{\our} Search Process.}
\label{alg:reward-update}
\small
\begin{algorithmic}[1]
\Require Root node $S_{root}$, number of rollouts $\mathcal{M}$, exploration constant $\epsilon$, memory thresholds $\tau_{pos},\tau_{low}$
\Ensure Updated search tree statistics $V,N$, memory states $\mathcal{MEM}_H, \mathcal{MEM}_F$

\State Initialize $\mathcal{MEM}_H \gets \emptyset, \mathcal{MEM}_F \gets \emptyset$
\For{$j = 1$ to $\mathcal{M}$}
    \State $C \gets S_{root}$
    \State -------------------------Selection-------------------------
    \While{$C$ is not leaf node}
    \State $C \gets argmax_{C' \in children(C)} \left( \mathcal{V}_{C'} + \epsilon \sqrt{\frac{2 \ln N_{C}}{N_{C'}}} \right)$
    \EndWhile

    \State -------------------------Expansion-------------------------
    \If{$C$ is not terminal}
    \State $C^{*} \gets FindChilds(C, \mathcal{MEM}_F, \mathcal{MEM}_H)$
    \For{each child $c$ in $C^{*}$}
        \State $Update(\mathcal{V}_c)$
    \EndFor
    \EndIf
    \State -------------------------Simulation-------------------------
    \While{$C$ is not terminal}
        \If{$C$ is not full Expanded}
        \State $Expand(C)$
        \EndIf
        \For{each $c$ in $children(C)$}
            \State \textbf{if} $\mathcal{V}_c \ge \tau_{pos}$ \textbf{then} Add $c$ to $\mathcal{MEM}_H$
            \State \textbf{if} $\mathcal{V}_c \le \tau_{neg}$ \textbf{then} Add $c$ to $\mathcal{MEM}_F$
        \EndFor
        \State $C \gets argmax_{C' \in children(C)}(\mathcal{V}_{C'})$
    \EndWhile
    \State ---------------------Backpropagation---------------------
    \State $UpdateStatistics(N_{C'})$ \Comment{Update all path nodes}
\EndFor

\Return Search Tree
\end{algorithmic}
\end{algorithm}

\begin{figure*}[htbp]
\centering 
\includegraphics[width=0.98\textwidth]{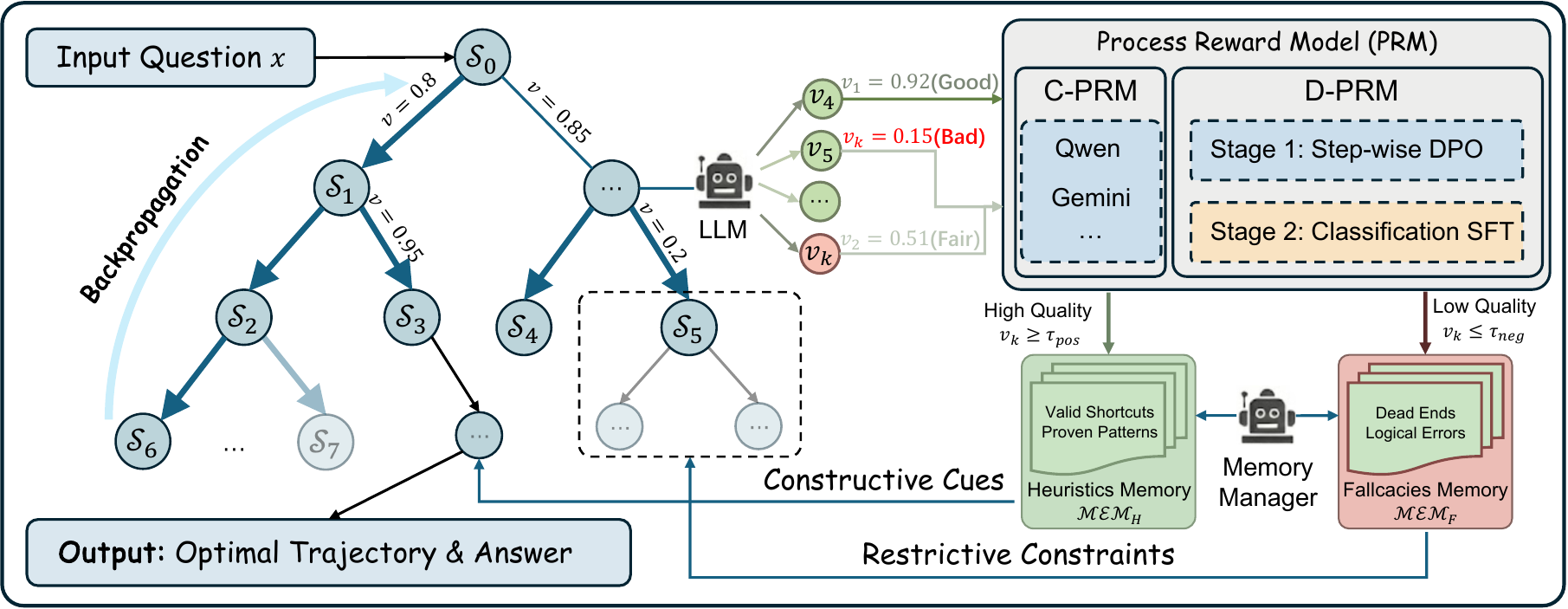}
\caption{Overview of the {\our} framework. It integrates MCTS with a PRM (Continuous-PRM or Discrete-PRM) and a reflective memory system. Stored \textbf{Heuristics Memory} and \textbf{Fallacies Memory} facilitate global information sharing to guide generation and prune the search space efficiently.}
\label{fig:main}
\end{figure*}

\paragraph{Problem Formulation.}
We define the task as generating a sequence of steps to solve a given problem $x$. 
The solution is represented as a trajectory $p_T = [s_1, s_2, \dots, s_T]$, where each $s_t$ denotes a reasoning step and $s_T$ includes the final answer. 
The objective is to identify the optimal trajectory that maximizes the likelihood of correctness. 

\subsection{Human-Like Reflective Memory}
{\our} integrates a memory mechanism inspired by human reflection. 
As shown in Figure~\ref{fig:main}, this system comprises three essential components: \textbf{Heuristics Memory}, \textbf{Fallacies Memory}, and a \textbf{Memory Manager}.
The overall search procedure is summarized in Algorithm~\ref{alg:reward-update}.

\paragraph{Heuristics Memory.} The Heuristics module ($\mathcal{MEM}_H$) archives optimal reasoning trajectories and validated intermediate states. 
During MCTS simulation, if a node's evaluated value $v_k$ exceeds a predefined positive threshold $\tau_{pos}$, its content is integrated into $\mathcal{MEM}_H$. 
These stored insights serve as guiding signals for subsequent expansions to facilitate the replication of successful strategies.

\paragraph{Fallacies Memory.} The Fallacies module ($\mathcal{MEM}_F$) serves as a repository for erroneous reasoning patterns and failed trajectories. 
When a node’s value $v_k$ falls below a negative threshold $\tau_{neg}$, it is archived within $\mathcal{MEM}_F$ to provide negative feedback. 
This mechanism provides essential constraints for pruning the search space, enabling the model to circumvent recurrent logical errors and improve search throughput.

\paragraph{Memory Manager.} The Memory Manager orchestrates interactions between search and memory modules to maintain informational parsimony.
Its primary function is to distill pivotal patterns while filtering out redundant entries, thereby preventing computational overhead. 
By ensuring that only high-fidelity insights remain in $\mathcal{MEM}_H$ and $\mathcal{MEM}_F$, the manager optimizes the quality of guidance provided during the reasoning process.

\subsection{Dual-Stage Process Reward Model}
Inspired by AR-MCTS~\cite{zhang2025ARMCTS}, we adopt a two-stage training strategy for our PRM. 
However, unlike previous approaches that rely on large-scale datasets, our method is tailored for \textbf{few-shot} scenarios where data scarcity is a challenge. 
To address this, we introduce a classification-based objective in the second stage.

\paragraph{Data Construction.}
We adopt the self-training pipeline from ReST-MCTS$^{*}$~\cite{zhang2024restmcts} to construct high-quality data.
Based on the search tree generated by MCTS, we calculate a quality value $v_k$ for each partial solution $p_k = [s_1, \dots, s_k]$, which serves as the ground truth.
The value $v_k$ is iteratively updated to reflect the cumulative progress towards the correct answer:
\begin{equation}
v_k = \max(v_{k-1} + w_{s_k}, 0)
\end{equation}
where $w_{s_k}$ is the weighted reward for step $s_k$.
To ensure the value reflects both step correctness and reasoning progress, $w_{s_k}$ incorporates the reasoning distance $m_k$ (minimum steps to the correct answer) and a step-level error indicator $r_{s_k}$:
\begin{equation}
w_{s_k} = \frac{1 - v_{k-1}}{m_k + 1} (1 - 2r_{s_k})
\end{equation}
Here, $r_{s_k}$ is derived from the search tree (set to 0 if the step is on a correct path, 1 otherwise).

\paragraph{Preference Alignment via SDPO.}
In the first stage, we align the model's preferences using Step-level Direct Preference Optimization (SDPO). 
By leveraging the reasoning paths generated during MCTS, we construct preference pairs $(y^+, y^-)$, where $y^+$ represents a high-quality step ($v_k \ge 0.8$) and $y^-$ represents a low-quality or incorrect step ($v_k \le 0.2$).  
The training objective is to maximize the likelihood of the preferred step over the dispreferred one relative to a reference model $\pi_{\text{ref}}$:
\begin{equation}
\begin{aligned}
\mathcal{L}_{\text{SDPO}}&(\pi_\theta; \pi_{\text{ref}}) = -\mathbb{E}_{(s_i, y^+, y^-) \sim \mathcal{D}} \Big[ \log \sigma \Big( \\
&\beta \log \frac{\pi_\theta(y^+|s_i)}{\pi_{\text{ref}}(y^+|s_i)} - \beta \log \frac{\pi_\theta(y^-|s_i)}{\pi_{\text{ref}}(y^-|s_i)} \Big) \Big]
\end{aligned}
\end{equation}
where $\beta$ is a hyperparameter controlling the deviation from the reference model.

\paragraph{Fine-Grained Value Classification.}
In the second stage, we reframe value estimation as classification rather than regression. 
This shift is motivated by the intuition that discrete categories generalize better than continuous values in few-shot settings.
Specifically, we discretize the original scores $v_k \in [0, 1]$ into five categories $\mathcal{C} = \{\text{Perfect, Good, Fair, Poor, Bad}\}$ using a uniform interval of $0.2$. 
Given a step $s_i$ and its ground truth label $y_i \in \mathcal{C}$, we optimize the cross-entropy loss:
\begin{equation}
\mathcal{L}_{\text{CLS}} = - \sum_{c \in \mathcal{C}} \mathbb{I}(y_i = c) \log P_\theta(c | s_i)
\end{equation}
where $P_\theta(c | s_i)$ is the probability assigned to class $c$ by the PRM. 

\begin{table*}[t] 
    \centering 
    \footnotesize
    
    \setlength{\tabcolsep}{10pt}{
        \begin{tabular}{l l c c c c c c}
            \toprule
            \multirow{2}{*}[-1pt]{Model} & \multirow{2}{*}[-1pt]{Method} & \multicolumn{2}{c}{\textsc{\textbf{GPQA-Diamond}}} & \multicolumn{2}{c}{\textsc{\textbf{FMT}}} & \multicolumn{1}{c}{\textsc{\textbf{MATH500}}} & \multicolumn{1}{c}{\textsc{\textbf{AIME25}}} \\
            \cmidrule(r){3-4} \cmidrule(r){5-6} \cmidrule(r){7-7} \cmidrule(r){8-8}
            & & EM $\uparrow$ & F1 $\uparrow$ & EM $\uparrow$ & F1 $\uparrow$ & Score $\uparrow$ & Score $\uparrow$ \\
            \cmidrule{1-8}
            \multirow{6}*{GPT-4.1-mini}
            & Zero-shot & 47.55 & 46.55 & 68.34 & 68.75 & 79.25 & 48.67 \\
            & ReAct & 55.05 & 60.75 & 65.00 & 67.61 & 79.10 & 50.00 \\
            & Search-O1 & 57.07 & 57.16 & 66.00 & 65.03 & 80.59 & 43.33 \\
            & ReST-MCTS$^{*}$ & 60.61 & 60.53 & 69.00 & 68.91 & 70.90 & 20.00 \\
            & MCTS-RAG & 64.65 & 64.80 & 69.50 & 68.98 & 80.60 & 50.00 \\
            & \cellcolor{gray!15}\our & \cellcolor{gray!15}\textbf{65.08} & \cellcolor{gray!15}\textbf{65.16} & \cellcolor{gray!15}\textbf{70.50} & \cellcolor{gray!15}\textbf{70.07} & \cellcolor{gray!15}\textbf{82.09} & \cellcolor{gray!15}\textbf{53.33} \\
            \midrule
            \multirow{6}{2.1cm}{Qwen3-30B-A3B-2507} 
            & Zero-shot & 45.80 & 45.29 & 62.72 & 62.85 & 75.07 & 53.33 \\
            & ReAct & 48.99 & 53.49 & 65.50 & 65.43 & 70.90 & 50.00 \\
            & Search-O1 & 61.11 & 53.76 & 68.00 & 67.76 & 88.06 & 50.00 \\
            & ReST-MCTS$^{*}$ & 58.08 & 60.45 & 64.00 & 63.43 & 77.61 & 20.00 \\
            & MCTS-RAG & 63.64 & 64.26 & \textbf{70.50} & \textbf{69.99} & \textbf{94.03} & \textbf{66.67} \\
            & \cellcolor{gray!15}\our & \cellcolor{gray!15}\textbf{65.15} & \cellcolor{gray!15}\textbf{66.26} & \cellcolor{gray!15}68.00 & \cellcolor{gray!15}67.41 & \cellcolor{gray!15}93.28 & \cellcolor{gray!15}63.33 \\

            \bottomrule
        \end{tabular}
    }
    \caption{Main performance comparison of {\our} against various baselines on scientific verification (GPQA-Diamond, FMT) and mathematical reasoning (MATH, AIME25) benchmarks. Best results for each backbone model are highlighted in \textbf{bold}.}
    \label{tab:main_result}
\end{table*}

\section{Experiments}
\subsection{Benchmarks and Metrics}
To evaluate the effectiveness of {\our}, we selected two types of benchmarks designed to stress-test the model’s performance across both knowledge depth and logical complexity:
\paragraph{Science and Fact Verification.}
This category assesses knowledge-intensive processing and information discernment via: (1) GPQA Diamond~\cite{Rein2023GPQA}, which focuses on high-level scientific question answering; and (2) FoolMeTwice (FMT)~\cite{Eisenschlos2023fmt}, a challenging fact-checking benchmark that requires the verification of complex factual claims.

\paragraph{Mathematical Reasoning.}
This category evaluates the model's capacity for rigorous logical deduction via: (1) MATH500~\cite{Lightman2024prm}, where we specifically select the most challenging problems (Level 5) to test multi-step problem-solving across competition-level mathematics; and (2) AIME25~\cite{aime25}, which utilizes problems from the American Invitational Mathematics Examination to challenge the model's advanced reasoning limits.

\paragraph{Evaluation Metrics.}
For Science and Fact Verification, we adopt two widely used metrics: Exact Match (EM), and F1 scores. 
For Mathematical Reasoning, performance is measured by Accuracy (Score), with MCTS-based methods deriving the final answer by selecting the maximum probability.
Furthermore, to validate the efficiency of MCTs, we introduce two structural metrics:
\textbf{Trajectory:} Quantifies the expansion breadth of the search tree to assess exploration range.
\textbf{Depth:} Measures the expansion layers to evaluate the length and complexity of the reasoning chains. 

\subsection{Baseline Systems}
To assess the efficacy of {\our}, we benchmark it against a set of frontier reasoning baselines, spanning from prompting techniques to sophisticated agentic and search-based frameworks.
\paragraph{Zero-Shot CoT.}
We employ standard Zero-Shot CoT prompting to elicit reasoning capabilities from the LLMs without task-specific examples. 
\paragraph{ReAct.}
ReAct~\cite{yao2023react} synergizes reasoning and acting by interleaving the generation of reasoning traces and task-specific actions. 
This allows the model to dynamically retrieve external information to support its reasoning process, enabling it to refine its understanding based on external evidence.
\paragraph{Search-o1.}
Search-o1~\cite{search_o1} enhances large reasoning models with an agentic RAG mechanism. 
It integrates a ``Reason-in-Documents'' module that refines retrieved documents into concise reasoning steps, allowing the model to autonomously retrieve and integrate external knowledge on demand while maintaining reasoning coherence.
\paragraph{Rest-MCTs.}
Rest-MCTs~\citep{zhang2024restmcts} introduces a process reward model (PRM) guided tree search policy. 
It utilizes a self-training pipeline where the policy model and PRM are iteratively improved using high-quality traces generated via MCTS. 
This method focuses on enhancing internal reasoning capabilities through search and self-improvement.
\paragraph{MCTS-RAG.}
MCTS-RAG~\cite{Hu2025MCTsRAG} integrates Monte Carlo Tree Search with Retrieval-Augmented Generation. 
It expands the action space of standard MCTS to include retrieval-specific actions, enabling the model to explore multiple reasoning and retrieval trajectories. 
This allows for dynamic knowledge acquisition and the selection of the most consistent reasoning path through voting mechanisms.

To ensure fair comparisons, we use the same base LLMs: {\qwen} and GPT-4.1-mini for all evaluated systems.
We keep all hyperparameters consistent (e.g., $\epsilon$, $\tau_{pos}$, $\tau_{neg}$).

\subsection{Main Results}
\paragraph{Performance on Science and Fact Verification.}
As shown in Table~\ref{tab:main_result}.
In knowledge-intensive tasks, {\our} exhibits remarkable robustness. 
On the challenging GPQA-Diamond benchmark, {\our} achieves state-of-the-art results. 
With GPT-4.1-mini, {\our} reaches an EM score of 65.08\%, significantly outperforming the standard Zero-shot baseline and the strong retrieval-augmented baseline MCTS-RAG. 
Similarly, using the {\qwen} backbone, {\our} achieves the highest EM of 65.15\%, surpassing Search-O1 and MCTS-RAG.
On the FMT, {\our} with GPT-4.1-mini achieves the best performance with an EM of 70.50\% and F1 of 70.07\%, effectively mitigating hallucination risks common in direct reasoning. 
While MCTS-RAG shows slightly higher scores on FMT with {\qwen}, {\our} remains highly competitive and significantly outperforms Zero-shot and ReAct baselines. 
However, the primary advantage of {\our} lies in its search efficiency(\S \ref{search_efficiency}).

\paragraph{Performance on Mathematical Reasoning.}
In the domain of logical deduction, {\our} demonstrates strong reasoning capabilities. 
On MATH500 and AIME25, {\our} with GPT-4.1-mini secures the top position, achieving scores of 82.09 and 53.33, respectively, surpassing both ReAct and MCTS-RAG. 
Notably, on the AIME25 benchmark, which tests advanced reasoning limits, {\our} shows a clear advantage over Search-O1.
For the {\qwen}, {\our} maintains high performance, comparable to the best-performing baseline MCTS-RAG and significantly exceeding Zero-shot and Search-O1. 
This indicates that integrating MCTS with our proposed mechanism enhances the model's ability to navigate complex solution spaces in mathematical problems.

Across all tasks, {\our} shows consistent improvements over the Zero-shot and ReAct baselines, validating the efficacy of integrating MCTS-based planning with retrieval augmentation. 
The substantial gains in GPQA and the competitive results in math benchmarks highlight the versatility of {\our} in handling both knowledge-retrieval heavy and logic-heavy tasks.

\begin{figure}[htbp]
    \centering
    \includegraphics[width=0.48\textwidth]{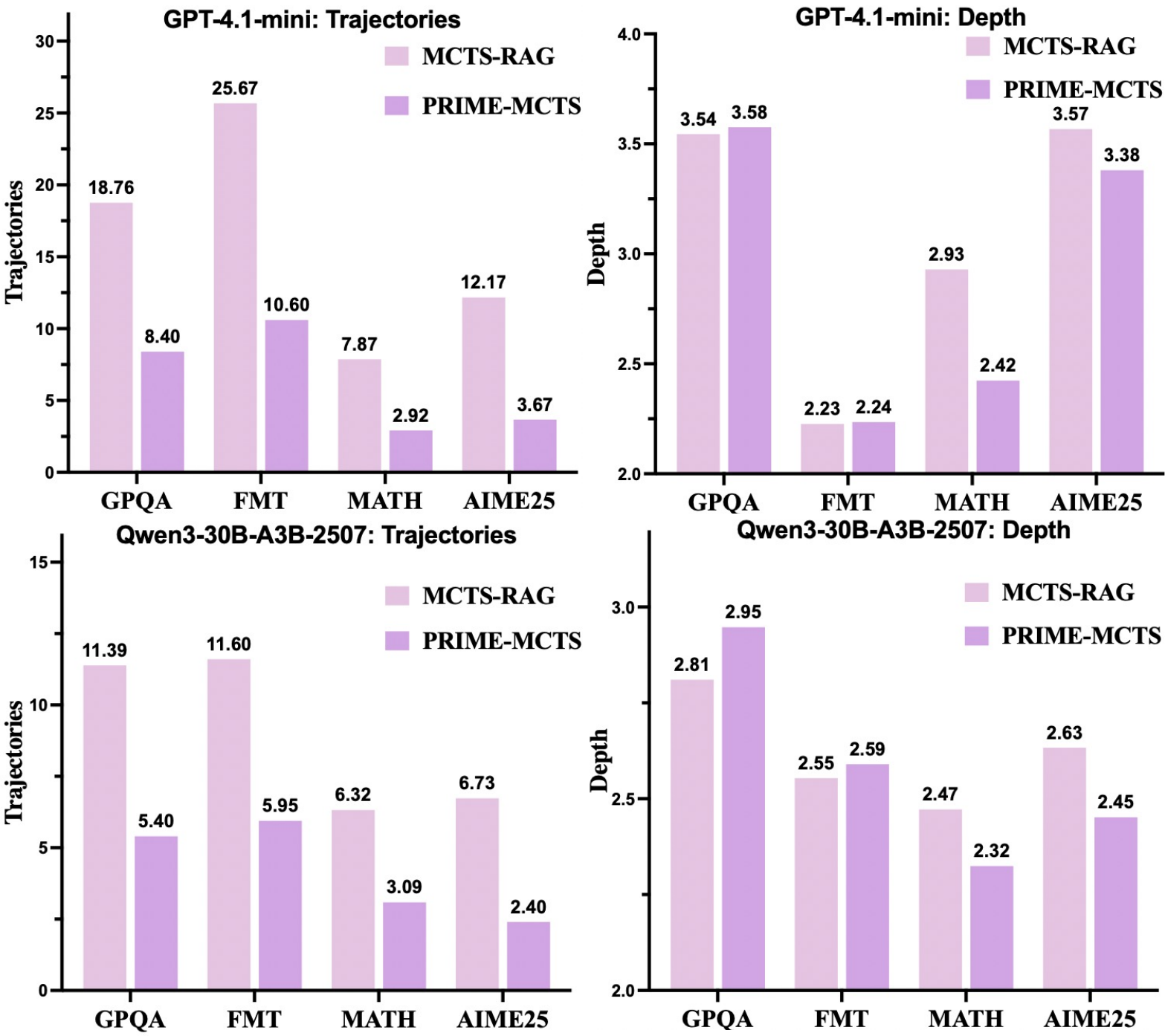}
    \caption{
    Comparison of search efficiency between MCTS-RAG and {\our}. The left panels display the average number of explored Trajectories (search breadth), while the right panels show the average reasoning Depth. {\our} consistently reduces the search space while maintaining optimized reasoning paths.}
    \label{fig:efficiency}
\end{figure}

\subsection{Search Efficiency Analysis}
\label{search_efficiency}
To further assess the computational efficiency of {\our}, we analyze the structural characteristics of the generated search trees compared to MCTS-RAG. 
Specifically, we focus on two key metrics: \textbf{Trajectories}, which measures the breadth of exploration (i.e., the number of valid reasoning paths explored), and \textbf{Depth}, which indicates the average length of the reasoning chains. 
Figure~\ref{fig:efficiency} illustrates these statistics across four benchmarks using both GPT-4.1-mini and {\qwen} as backbones.

\paragraph{Reduced Search Trajectories.}
As shown in the left panels of Figure~\ref{fig:efficiency}, {\our} demonstrates a significant reduction in the number of trajectories compared to MCTS-RAG across all datasets. 
For instance, on the GPQA benchmark with GPT-4.1-mini, {\our} reduces the average number of trajectories from 18.76 to 8.40, a decrease of over 55\%. 
Similarly, on the AIME25 dataset with {\qwen}, the trajectories drop from 6.73 to 2.40. 
This substantial reduction indicates that {\our} effectively prunes the search space. 
By leveraging the Memory mechanism through Heuristics ($\mathcal{MEM}_H$) and Fallacies ($\mathcal{MEM}_F$), the model identifies and prunes suboptimal or redundant trajectories during the search process.
This allows {\our} to concentrate its computational resources on the most plausible reasoning directions instead of brute-force tree expansion.

\paragraph{Optimized Reasoning Depth.}
Regarding reasoning depth, {\our} maintains a comparable or slightly optimized depth relative to MCTS-RAG. 
On knowledge-intensive tasks like GPQA and FMT, the depth remains stable, ensuring that the reasoning rigor is preserved. 
In mathematical tasks like MATH500 and AIME25, we observe a slight reduction in depth, suggesting that the memory-augmented guidance helps the model find more direct and efficient solutions without unnecessary intermediate steps.

\paragraph{Conclusion.}
The combination of significantly reduced search breadth and stable reasoning depth confirms the efficiency of {\our}. It achieves superior performance (as shown in Table~\ref{tab:main_result}) while consuming fewer search steps, demonstrating that the integration of parametric memory into the MCTS framework guides the model to reason ``smarter'' rather than just ``harder.''

\begin{table}[t]
\centering
\small
\resizebox{1.0\columnwidth}{!}{
\resizebox{0.5\textwidth}{!}{
\begin{tabular}{l | ccccc}
\toprule
  & EM$\uparrow$ & F1$\uparrow$ & LA$\uparrow$ & Traj$\downarrow$ & Depth$\downarrow$ \\
\midrule
\multicolumn{6}{c}{\textbf{GPT-4.1-mini}} \\
\midrule
Oracle-PRM & 70.50 & \textbf{70.07} & 76.00 & \textbf{10.60} & \textbf{2.24} \\
Light-PRM    &  \textbf{70.50} & 69.91 & \textbf{80.50} & 14.91 & 2.33 \\
\midrule
\multicolumn{6}{c}{\textbf{Qwen3-30B-A3B-2507}} \\
\midrule
Oracle-PRM & 68.00 & 67.41 & \textbf{82.50} & \textbf{5.95} & 2.59 \\
Light-PRM  & \textbf{68.00} & \textbf{67.41} & 79.50 & 6.65 & \textbf{2.52} \\
\bottomrule
\end{tabular}
}%
}
\caption{Comparison of reasoning performance and search efficiency between the frontier Gemini-2.5-Pro and our proposed locally-trained dual-stage PRM.}
\label{tab:comp_prm}%
\end{table}

\subsection{Dual-Stage Process Reward Model}
To evaluate the effectiveness of our two-stage training strategy (SDPO and Fine-Grained Classification), we investigate whether the distilled local Dual-Stage Process Reward Model can match the performance of high-capability closed-source models. 
Specifically, we compare two configurations of the reward evaluator:
(1) \textbf{Oracle-PRM}: Utilizing Gemini-2.5-Pro~\cite{Gemini2025} to provide reward scoring and memory management, representing a high-performance upper bound.
(2) \textbf{Light-PRM}: Employing our locally fine-tuned Qwen-3-4B model, trained via the two-stage pipeline, to perform these auxiliary roles.
Table~\ref{tab:comp_prm} presents the comparative results using both GPT-4.1-mini and {\qwen} as the core reasoning backbones.

\paragraph{Reward Capability Alignment.}
Remarkably, the \textbf{Light-PRM} achieves performance parity with the \textbf{Oracle}-guided system. 
With the GPT-4.1-mini backbone, the distilled model matches the oracle exactly in Exact Match (EM) at 70.50\% and maintains a highly competitive F1 score (69.91 vs. 70.07). 
Similarly, under the {\qwen} backbone, the local model achieves identical EM (68.00\%) and F1 (67.41\%). 
These results demonstrate that our proposed self-training pipeline effectively distills the critical discrimination and planning capabilities, such as identifying valid reasoning steps and potential pitfalls.

\begin{figure}[htbp]
    \centering
    \includegraphics[width=0.48\textwidth]{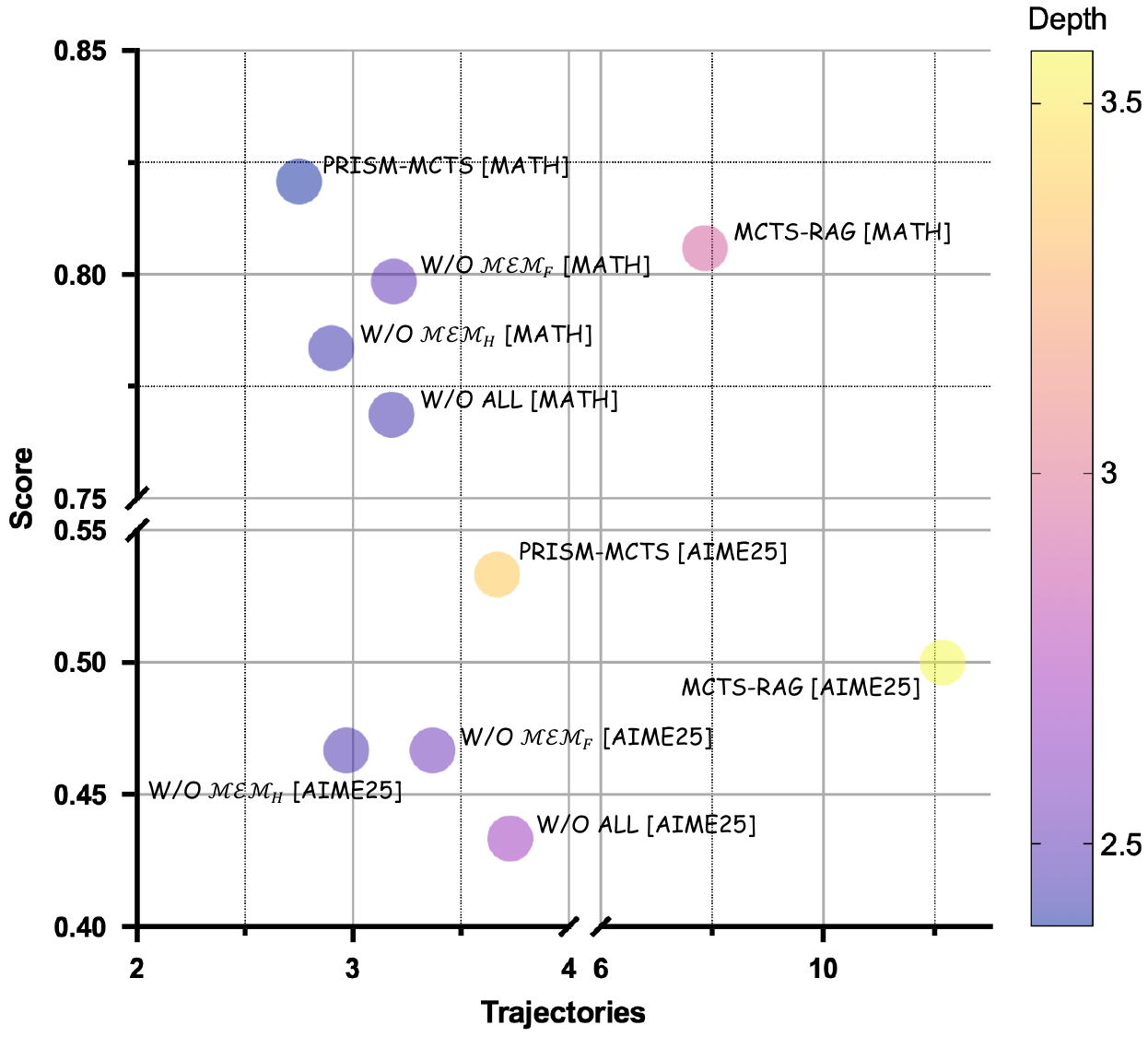}
    \caption{Ablation study on MATH500 and AIME25 benchmarks evaluating the impact of Heuristics Memory ($\mathcal{MEM}_H$) and Fallacies Memory ($\mathcal{MEM}_F$) modules. The scatter plot correlates search efficiency (Trajectories) with reasoning performance (Score), where color indicates reasoning Depth. Removing memory components leads to performance degradation, while the full {\our} maintains high accuracy with minimal search breadth, significantly outperforming the MCTS-RAG baseline in efficiency.}
    \label{fig:ablation}
\end{figure}

\paragraph{Efficiency Analysis.}
In terms of search efficiency, the Local model exhibits a slightly larger search breadth compared to the Gemini oracle. 
As shown in the \textit{Trajectories} column of Table~\ref{tab:comp_prm}, the average number of trajectories for the Local model is higher. 
This suggests that the local PRM is slightly less discriminative in the early stages of search, necessitating a broader exploration to locate the optimal solution. 
However, the reasoning \textit{Depth} remains stable, confirming that the local guidance successfully prevents the reasoning from degenerating into inefficient, overly long chains.
The minimal performance gap between the Local and Gemini configurations highlights the robustness of \our. 
It offers a flexible trade-off: users can leverage powerful APIs for maximum efficiency, or opt for the local deployment to achieve efficient reasoning capabilities with full data privacy and reduced operational costs.

\subsection{Ablation Studies}
We perform ablation studies on MATH500 and AIME25 benchmarks to evaluate the specific contributions of $\mathcal{MEM}_H$ and $\mathcal{MEM}_F$. We compare the full {\our} model against three configurations: (1) w/o $\mathcal{MEM}_H$, where the Heuristics module is disabled; (2) w/o $\mathcal{MEM}_H$, where the Fallacies module is removed; and (3) w/o ALL, where both memory components are deactivated.

Figure~\ref{fig:ablation} illustrates the results, where the $x$-axis represents the search breadth (Trajectories), the $y$-axis represents the model performance (Score), and the marker color corresponds to the reasoning Depth (as shown in the color bar).

\paragraph{Impact of Memory on Efficiency and Accuracy.}
The most striking observation is the clustering of PRISM-MCTS and its ablation variants on the left side of the plot, maintaining high performance with very few trajectories ($\approx 3$). 
In contrast, the MCTS-RAG baseline relies on a significantly broader search, requiring roughly $2\times$ to $4\times$ more trajectories to achieve comparable or even lower scores.
In contrast, {\our} achieves higher (or comparable) scores with substantially fewer trajectories.
However, when memory is fully removed, the performance drops drastically (e.g., on MATH500, from $\sim$0.82 to $\sim$0.77).
This finding demonstrates that the efficiency of {\our} is not a mere byproduct of breadth restriction; instead, the Memory mechanism provides the critical navigational guidance required to pinpoint correct solutions within a compact search space. 

\paragraph{Relative Importance of Heuristics and Fallacies.}
We observe that both Heuristics Memory and Fallacies Memory are essential for peak performance, though their relative impact varies. 
On the MATH500 benchmark, deactivating Heuristics (w/o $\mathcal{MEM}_H$) causes a more pronounced decline than removing Fallacies (w/o $\mathcal{MEM}_F$). 
This suggests that positive reinforcement through successful trajectories is marginally more critical for mathematical reasoning than error avoidance. 
Ultimately, the synergy between these modules yields the most robust reasoning policy by balancing constructive guidance with negative constraints.

\section{Conclusion}

In this paper, we present {\our}, a framework that advances inference-time scaling by integrating a PRM with a reflective shared memory system.
By dynamically leveraging \textit{Heuristics Memory} and \textit{Fallacies Memory}, {\our} delivers strong performance on complex benchmarks while substantially reducing the search breadth.
Furthermore, we propose a resource-efficient PRM training strategy enabling local models to rival the fidelity of top closed-source PRM model.
Overall, our method demonstrates that through memory reflection, we can effectively enhance MCTS search efficiency and achieve higher-quality reasoning.

\section*{Limitations}

\paragraph{Limited PRM Training Scale.}
Due to resource constraints, the volume of synthesized training data for the Process Reward Model was restricted.
We hypothesize that if sufficient computational resources were available to generate a larger corpus of high-quality process supervision data, the model would demonstrate superior domain adaptation capabilities, thereby leading to more robust reasoning performance.

\paragraph{Absence of Multi-modal Support.}
Our current implementation and evaluation of {\our} are primarily confined to text-based reasoning tasks.
While the proposed reflective memory mechanism and PRM-guided search framework are theoretically applicable to multi-modal contexts (e.g., visual reasoning or diagram-based mathematical problem solving), the current system has not yet been extended to handle non-textual inputs.
As advanced reasoning tasks increasingly require the integration of cross-modal information, the lack of multi-modal capabilities remains a limitation of the current work.


\bibliography{custom}




\end{document}